# Learning the Bayesian Network Structure: Dirichlet Prior versus Data

**Harald Steck**
Computer-Aided Diagnosis (IKM CKS)
Siemens Medical Solutions, 51 Valley Stream Parkway E51
Malvern, PA 19355, USA
*harald.steck@siemens.com*

## Abstract

In the Bayesian approach to structure learning of graphical models, the equivalent sample size (ESS) in the Dirichlet prior over the model parameters was recently shown to have an important effect on the maximum-a-posteriori estimate of the Bayesian network structure. In our first contribution, we theoretically analyze the case of *large* ESS-values, which complements previous work: among other results, we find that the presence of an edge in a Bayesian network is favoured over its absence even if both the Dirichlet prior and the data imply independence, as long as the conditional empirical distribution is notably different from uniform. In our second contribution, we focus on realistic ESS-values, and provide an analytical approximation to the 'optimal' ESS-value in a predictive sense (its accuracy is also validated experimentally): this approximation provides an understanding as to which properties of the data have the main effect determining the 'optimal' ESS-value.

## 1 INTRODUCTION

The posterior probability and the marginal likelihood of the graph are one of the most popular scoring functions for learning Bayesian network structures, e.g. [Buntine, 1991, Heckerman *et al.*, 1995]. Both of them require the value of a free parameter to be specified by the researcher: the so-called *equivalent sample size* (ESS), which originates from the Dirichlet prior over the model parameters, cf. [Buntine, 1991, Heckerman *et al.*, 1995]. In the elaborate experiments on 20 UCI data sets in [Silander *et al.*, 2007], it was shown that the chosen ESS-value has a decisive effect on the resulting maximum-a-posteriori (MAP) graph estimate: in the experiment shown in Figure 1 in [Silander *et al.*, 2007], the number of edges increases monotonically from zero to the maximum as the ESS-value grows. These two extremes (empty and complete graph) are indeed reached for several data sets, and they are almost reached for the remaining data sets in [Silander *et al.*, 2007].

The effect of the ESS-value on the MAP graph estimate was earlier noticed in [Steck & Jaakkola, 2002], and one of its main contributions was the theoretical analysis of the case of *small* ESS-values. That paper also provided a result for the case ESS→ ∞: the Bayes factor converges to zero *in the limit*, which means that it favours neither the presence nor the absence of an edge. Given this inconclusive behavior *in the limit*, our first contribution in this paper is concerned with the case of *large but finite* ESS-values: in our theoretical analysis in Section 3, we derive the properties of the given data under which the presence of an edge in the graph is favoured. This contribution concerning the case of large but finite ESS-values, combined with the results for small ESS-values [Steck & Jaakkola, 2002], implies immediately that, under these conditions, the number of edges increases when the ESS-value grows, although not necessarily monotonically. Our second contribution is concerned with the case of *realistic / intermediate* ESS-values (Section 4): our goal is to *understand* as to which properties of a given data set determine the 'optimal' ESS-value (in the predictive sense). We achieve this by deriving an explicit analytical approximation (its validity is also assessed experimentally).

## 2 BRIEF REVIEW OF BAYESIAN NETWORK LEARNING

A Bayesian network model comprises a directed acyclic graph $G$, and the model parameters $\theta$ are conditional probabilities, e.g., see [Cowell *et al.*, 1999] for an overview. It describes a joint probability distribution $p(X|G)$ over the vector of random variables $X = (X_1, ..., X_n)$, where $n$ is the number of variables. We consider discrete random variables $X_i$ with a multinomial distribution in this paper. According to the graph $G$, the joint distribution factors recursively, $p(X|G) = \prod_{i=1}^n \theta_{X_i|\Pi_i}$; $\Pi_i$ denotes the set of parents of variable/node $X_i$ in the graph $G$; the (joint) states of $\Pi_i$ will be denoted by $\pi_i$, and the states of $X_i$ by $x_i$.

Various approaches to learning the graph $G$ from given data $D$ have been developed in the literature, e.g., see [Cowell *et al.*, 1999] for an overview. The Bayesian approach appears to be most popular. As a scoring function, it employs the posterior probability of the graph $G$ given the data, $p(G|D) \propto p(D|G)p(G)$, or the marginal likelihood of the graph, $p(D|G)$. Both are equivalent if the prior distribution over the graphs, $p(G)$, is chosen to be uniform. The marginal likelihood of the graph can be expressed in closed form when using the Dirichlet prior over the model parameters [Buntine, 1991, Heckerman *et al.*, 1995]. More importantly, the Dirichlet prior is the only distribution that ensures likelihood equivalence [Heckerman *et al.*, 1995], a desirable property in structure learning. In particular, likelihood equivalence holds only if the hyper-parameters $\alpha_{x_i,\pi_i}$ of the Dirichlet prior can be expressed as $\alpha_{x_i,\pi_i} = \alpha \cdot q_{x_i,\pi_i}$ for all $x_i, \pi_i$ and for all $i = 1,...,n$; $q$ is a prior distribution over $X$, and $\alpha$ is a positive constant independent of $i$, the so-called scale-parameter or *equivalent sample size* (ESS). When $q$ is chosen to be a uniform distribution, one arrives at the BDeu score [Buntine, 1991]: $q_{x_i,\pi_i} = 1/(|X_i| \cdot |\Pi_i|)$, where $|\cdot|$ denotes the number of (joint) states of the variable(s). Hence,

$$\alpha_{x_i,\pi_i} = \frac{\alpha}{|X_i| \cdot |\Pi_i|}, \quad (1)$$

so that the ESS $\alpha$ is the only remaining free parameter in the marginal likelihood of the graph $G$: $p(D|\alpha, G)$ equals [Buntine, 1991, Heckerman *et al.*, 1995]

$$\prod_{i=1}^{n} \prod_{\pi_i} \frac{\Gamma(\alpha_{\pi_i})}{\Gamma(N_{\pi_i} + \alpha_{\pi_i})} \prod_{x_i} \frac{\Gamma(N_{x_i,\pi_i} + \alpha_{x_i,\pi_i})}{\Gamma(\alpha_{x_i,\pi_i})}, \quad (2)$$

where $\Gamma$ is the Gamma function; the first product extends over all the variables, while the second one is over all the joint states $\pi_i$ of the parents $\Pi_i$ of variable $X_i$, and the third product is over all the states $x_i$ of $X_i$. The cell counts $N_{x_i,\pi_i}$ in the contingency tables serve as sufficient statistics; the sample size is $N = \sum_{x_i,\pi_i} N_{x_i,\pi_i}$. In this Bayesian approach with the Dirichlet prior, the regularized parameter estimates are [Buntine, 1991]

$$\bar{\theta}_{x_i|\pi_i} \equiv E_{p(\theta_{x_i|\pi_i}|D,G)}[\theta_{x_i|\pi_i}] = \frac{N_{x_i,\pi_i} + \alpha_{x_i,\pi_i}}{N_{\pi_i} + \alpha_{\pi_i}}, \quad (3)$$

which is the expectation with respect to the parameter's posterior distribution.

For fixed data $D$ and ESS $\alpha$, finding the maximum-a-posteriori (MAP) estimate of the graph $G$, or the maximum with respect to the marginal likelihood in Eq. 2, is an NP-complete problem [Chickering, 1996]. One thus has to resort to heuristic search strategies, like local search (e.g., [Heckerman *et al.*, 1995]), as to find a close-to-optimal graph. If the number of variables is not prohibitively large, the exact solution, i.e., the globally optimal graph $G$, can be found in reasonable computation-time due to recent advances in structure learning [Silander & Myllymäki, 2006, Koivisto & Sood, 2004].

Besides this popular Bayesian score, various other scoring functions have been devised for structure learning, including the Bayesian information criteria (BIC) [Schwarz, 1978, Haughton, 1988], the Akaike Information Criteria (AIC) [Akaike, 1973], and the Minimum Description Length (MDL) [Rissanen, 1978]. We will use the AIC in our second contribution in Section 4.2.

## 3 LARGE ESS-VALUE: ITS EFFECT ON THE LEARNED GRAPH

In this section we present our first contribution, namely as to how a *large but finite* value of the ESS affects the learned optimal Bayesian network structure. This complements the theoretical results for *small* ESS-values derived in [Steck & Jaakkola, 2002] and the experimental results for *'intermediate'* ESS-values obtained in [Silander *et al.*, 2007].

### 3.1 UNIFORMITY-MEASURE

In this section, we define a new 'uniformity' measure and discuss its properties. This measure (and its properties) determines the result of structure learning for large ESS values, as we will see in the next section.

**Definition:** *We define the uniformity-measure U of a conditional multivariate distribution $p(A,B|\Pi)$ over two random variables A and B given a set of variables $\Pi$, as follows:*

$$U(p(A,B|\Pi))$$
$$= \sum_{a,b,\pi} p(a,b,\pi)\Big(|A,B,\Pi| \cdot p(a,b,\pi) - |A,\Pi| \cdot p(a,\pi)$$
$$- |B,\Pi| \cdot p(b,\pi) + |\Pi| \cdot p(\pi)\Big), \quad (4)$$

*where $|A,B,\Pi|$ denotes the number of joint states.*

Obviously, this definition is equivalent to:

$$U(p(A,B|\Pi)) = |\Pi| \cdot \sum_{\pi} p(\pi)^2 \sum_{a,b} p(a,b|\pi)$$
$$\cdot \Big(|A,B| \cdot p(a,b|\pi) - |A| \cdot p(a|\pi) - |B| \cdot p(b|\pi) + 1\Big)$$
$$= |A,B,\Pi| \cdot \sum_{a,b,\pi} p(a,b,\pi)^2 - |A,\Pi| \cdot \sum_{a,\pi} p(a,\pi)^2$$
$$- |B,\Pi| \cdot \sum_{b,\pi} p(b,\pi)^2 + |\Pi| \cdot \sum_{\pi} p(\pi)^2, \quad (5)$$

where $U$ is rewritten in terms of conditional probabilities in the first line, and in terms of squared probabilities in the second line.

The interesting property of $U$ (in each representation) is that it is a weighted sum of four terms, where the weights are the number of (joint) states of the variables.

**Proposition 1:** *The measure $U$ has the following three basic properties:*

- *symmetry: $U(p(A,B|\Pi)) = U(p(B,A|\Pi))$*
- *non-negativity: $U(p(A,B|\Pi)) \geq 0$*
- *minimality: $U(p(A,B|\Pi)) = 0$ if and only if*
  - *(conditional) independence:*
    *$p(A,B|\Pi) = p(A|\Pi)p(B|\Pi)$*
  - *and, for each state $\pi$ with $p(\pi) > 0$, at least one of the marginal distributions is uniform: $p(A|\pi) = 1/|A|$ or $p(B|\pi) = 1/|B|$.*

Concerning the minimality, note that $U > 0$ if a state $\pi$ with $p(\pi) > 0$ exists such that neither one of $p(A|\pi)$ or $p(B|\pi)$ is uniform: this includes distributions where $A$ and $B$ are conditionally independent, but both distributions are skewed. In other words, $U$ is not necessarily equal to zero for independent variables.

**Proof:** The symmetry is obvious. As to proof the other two properties, we focus on the representation in terms of conditional probabilities in Eq. 5; obviously, if these properties hold for each state $\pi$, then they hold also for $U$, the sum. As the remainder of the proof is understood to be conditional on a state $\pi$ with $p(\pi) > 0$, we now omit $\pi$ from the notation. A necessary condition for minimization of $U$ under the normalization-constraint $\sum_{a,b} p(a,b) = 1$ (accounted for by introducing the Lagrange multiplier $\lambda$), is that all the first partial derivatives vanish, i.e., for *all* states $a, b$:

$$2|A,B|p(a,b) - 2|A|p(a) - 2|B|p(b) + \lambda = 0 \quad (6)$$

With the normalization constraint it follows immediately that $\lambda = 2$. It follows for the particular choice of considering the difference between each pair of these equations pertaining to the same state $b$, but different states $a$ and $a'$:

$$p(a,b) - p(a',b) = [p(a) - p(a')]/|B|$$

Hence, if $p(a) = p(a')$ for all $a,a'$, then $p(A)$ is uniform and hence $p(A,B) = p(A)p(B)$ with arbitrary $p(B)$. Otherwise, i.e., if $p(a) \neq p(a')$ for some $a,a'$, then $p(A,B) = p(A)p(B)$, where $p(B)$ has to be uniform, and $p(A)$ can be arbitrary. Conversely, it can be verified easily that such a distribution indeed fulfills the original condition in Eq. 6. Moreover, $U = 0$ for such a distribution. Finally, it can be shown that the second derivative is positive definite, which completes the proof. □

**Aside:** Besides the interesting fact that $U$ is a weighted sum where the number of states function as the weights, we also like to mention that each of the four terms themselves is related to well-known quantities: the squared probabilities in Eq. 5 suggest a relationship to the well-known Gini index, $H^G(p(X)) = 1 - \sum_x p_x^2$, which is used as an impurity measure when learning decision trees. It can also be viewed as a special case of the Tsallis entropy $H^T_\beta(p(X)) = (1 - \sum_x p_x^\beta)/(\beta - 1)$ with parameter $\beta = 2$ [Tsallis, 2000]. The Tsallis entropy is used in statistical physics, and can be understood as the leading-order Taylor expansion of the Renyi entropy, $H^R_\beta(p(X)) = (\log \sum_x p_x^\beta)/(1 - \beta)$, which is a generalization of the Shannon entropy. In the limit $\beta \to 1$, the Tsallis entropy and the Renyi entropy both coincide with the Shannon entropy.

### 3.2 LEADING-ORDER APPROXIMATION OF BAYES FACTOR

In this section, we theoretically analyze the behavior of the Bayesian approach to structure learning, as reviewed in Section 2, for the case of *large but finite* ESS-values.

As to determine the most likely graph structure, note that the marginal likelihood $p(D|\alpha, G^+)$ of a graph $G^+$ is important only *relative* to the marginal likelihood $p(D|\alpha, G^-)$ of a competing graph $G^-$. In particular, we consider two graphs in the following that are identical except for the edge $A \leftarrow B$, which is present in $G^+$ and absent in $G^-$. Let $\Pi$ denote the parents of $A$ in graph $G^-$; this implies that, in $G^+$, the parents of $A$ are $\Pi$ and $B$. The presence of the edge $A \leftarrow B$ given the parents $\Pi$ is favoured over its absence (i.e., graph $G^+$ over $G^-$) if the log Bayes factor $\log p(D|\alpha, G^+)/p(D|\alpha, G^-)$ is positive; and if negative, the absence of $A \leftarrow B$ is favoured. Given complete data (i.e., no missing data), the marginal likelihood factorizes (cf. Eq. 2), so that most terms in the Bayes factor cancel out, and it depends only on the variables $A$ and $B$ and the parents $\Pi$:

$$\log \frac{p(D|\alpha, G^+)}{p(D|\alpha, G^-)} = \sum_{a,b,\pi} \log \frac{\Gamma(N_{a,b,\pi} + \alpha_{a,b,\pi})}{\Gamma(\alpha_{a,b,\pi})}$$
$$- \sum_{a,\pi} \log \frac{\Gamma(N_{a,\pi} + \alpha_{a,\pi})}{\Gamma(\alpha_{a,\pi})} - \sum_{b,\pi} \log \frac{\Gamma(N_{b,\pi} + \alpha_{b,\pi})}{\Gamma(\alpha_{b,\pi})}$$
$$+ \sum_{\pi} \log \frac{\Gamma(N_\pi + \alpha_\pi)}{\Gamma(\alpha_\pi)} \quad (7)$$

Note that this Bayes factor is symmetric in $A$ and $B$, as expected for independence tests; the asymmetry of the edge $A \leftarrow B$ is caused by $\Pi$ being the parents of $A$ (rather than of $B$) in the graph. Now we can present our main result for large but finite ESS-values:

**Proposition 2:** *Given the BDeu score, the leading-order approximation of the Bayes factor of the two graphs $G^+$*

and $G^-$ defined above reads for large ESS-values ($\alpha \gg N$):

$$\log \frac{p(D|\alpha, G^+)}{p(D|\alpha, G^-)}$$
$$= \frac{N}{2\alpha} \left\{ N \cdot U(\hat{p}(A,B|\Pi)) - d_F \right\} + \mathcal{O}(N^2/\alpha^2) \quad (8)$$

where $U$ is the uniformity measure as defined above, $\hat{p}(A,B|\Pi)$ is the empirical distribution implied by the data, i.e., $\hat{p}(a,b|\pi) = N_{a,b,\pi}/N_\pi$, and $d_F = |\Pi|(|A|-1)(|B|-1)$ are the (well known) degrees of freedom.[1]

Before we give the proof at the end of this section, let us first discuss interesting insights that follow from this approximation:

First, note that if $U$ were replaced by the (conditional) mutual information $I$, then the term $N \cdot I - d_F$ would be identical to the Akaike Information Criteria (AIC) [Akaike, 1973]. This analogy suggests that, in Eq. 8, $d_F$ serves as a penalty for model complexity. In other words, $N \cdot U(\hat{p}) - d_F > 0$ means that $U(\hat{p})$ is *notably* (or significantly) larger than zero, while $N \cdot U(\hat{p}) - d_F < 0$ refers to $U(\hat{p})$ being not notably larger than zero. Given the properties of the new measure $U$ (as discussed in Section 3.1), it hence follows directly:

**Corollary 1:** *Given the BDeu score, there exists a value $\alpha^+ \in \mathbb{R}$ such that for all finite $\alpha > \alpha^+$ the presence of the edge $A \leftarrow B$ given the parents $\Pi$ is favoured over its absence if $N \cdot U(\hat{p}(A,B|\Pi)) - d_F > 0$, i.e., if the empirical distribution $\hat{p}$ implies*

- *a notable dependence between A and B given $\Pi$,*

- **or** *a notable skewness (i.e., non-uniformity) of both distributions $\hat{p}(A|\Pi)$ and $\hat{p}(B|\Pi)$. Note that A and B can be conditionally independent here.*

*Conversely, the absence of the edge $A \leftarrow B$ given the parents $\Pi$ is favoured for all large values $\alpha \gg N$ if $N \cdot U(\hat{p}(A,B|\Pi)) - d_F < 0$, i.e., if the empirical distribution implies that*

- *A and B are not notably dependent given $\Pi$,*

- **and** *for each state $\pi$ with $p(\pi) > 0$ there is one marginal distribution $\hat{p}(A|\pi)$ or $\hat{p}(B|\pi)$ that is not notably different from a uniform distribution.*

Second, given that this applies to any edge in the graph, it follows immediately that the *complete* graph achieves the largest marginal likelihood if a sufficiently large (but finite)

---

[1] Here, $d_F$ must not be corrected for zero-cell counts, as it originates from the prior in the BDeu score, cf. Eq. 10. Note that this is different from $d_G^{\text{eff}}$ in Section 4.2.

ESS-value is chosen, assuming that, for each variable conditioned on any set of parents, the data implies notable dependence or a sufficiently skewed distribution. Note, however, that the latter condition may not necessarily hold for large sets of parents when the given data set is *small*; this is because many zero-cell counts occur if the joint number of states of the variables is larger than the number of samples in the data.

**Proof of Proposition 2:** Each of the four terms in the Bayes factor in Eq. 7 takes the form $\sum_y \log(\Gamma(N_y + \alpha_y)/\Gamma(\alpha_y))$, where $y$ denotes a (joint) state $y$ of a set $Y$ of random variables. If $\alpha_y \gg N_y$, we obtain for each $y$ (using $\Gamma(z) = (z-1)!$ in the first line):

$$\log \frac{\Gamma(N_y + \alpha_y)}{\Gamma(\alpha_y)} = \sum_{k=1}^{N_y} \log(k - 1 + \alpha_y)$$
$$= \sum_{k=1}^{N_y} \log \alpha_y + \sum_{k=1}^{N_y} \log \frac{k-1+\alpha_y}{\alpha_y}$$
$$= N_y \log \alpha_y + \sum_{k=1}^{N_y} \frac{k-1}{\alpha_y} + \mathcal{O}(N_y^2/\alpha_y^2)$$
$$= N_y \log \alpha_y + \frac{N_y(N_y-1)}{2\alpha_y} + \mathcal{O}(N_y^2/\alpha_y^2). \quad (9)$$

Inserting this approximation into the log Bayes factor of Eq. 7, we obtain (note that $\alpha_{a,b,\pi} = \alpha/|A,B,\Pi|$ for the BDeu score, where $|\cdot|$ denotes the number of joint states of the random variables):

$$\log \frac{p(D|G^+)}{p(D|G^-)}$$
$$= \sum_{a,b,\pi} N_{a,b,\pi} \log \frac{\alpha_{a,b,\pi} \alpha_\pi}{\alpha_{b,\pi} \alpha_{a,\pi}}$$
$$+ \frac{1}{2\alpha} \sum_{a,b,\pi} N_{a,b,\pi} \big( |A,B,\Pi| \cdot N_{a,b,\pi} - |A,\Pi| \cdot N_{a,\pi}$$
$$- |B,\Pi| \cdot N_{b,\pi} + |\Pi| \cdot N_\pi \big)$$
$$- \frac{1}{2\alpha} \sum_{a,b,\pi} N_{a,b,\pi} \big( |A,B,\Pi| - |A,\Pi| - |B,\Pi| + |\Pi| \big)$$
$$+ \mathcal{O}(N^2/\alpha^2) \quad (10)$$

where the first term vanishes because $\alpha_{A,B,\Pi}$ is uniform for the BDeu score, the second term equals $N^2 \cdot U(\hat{p}(A,B|\Pi))/(2\alpha)$,[2] and the third term equals $d_F \cdot N/(2\alpha)$. □

### 3.3 ILLUSTRATION

This section provides a simple example that shows how the combination of a uniform (prior) distribution and a skewed empirical distribution can create dependence while each individual distribution implies independence. The key is that, in this Bayesian approach with the Dirichlet prior, the log Bayes factor is essentially based on a weighted *sum* of two distributions (see Section 2 and Eq. 3): the empirical distribution $\hat{p}$ (with weight $N$) and the uniform distribution $q$ of the 'virtual' data points due to the Dirichlet prior (with weight $\alpha$). Obviously, the *sum* of two distributions (where each implies independence) does not necessarily result in a distribution that implies independence: $N\hat{p}(A)\hat{p}(B) + \alpha q(A)q(B) \neq c \cdot p(A)p(B)$ for any distribution $p$ and any constant $c$ in general. The deviation from this equality is typically not statistically significant for uniform $q$ and skewed $\hat{p}$ as long as the ESS-value $\alpha$ is sufficiently small; but as the ESS-value $\alpha$ increases in this weighted sum, the deviation can become statistically significant, as illustrated in the following example: let us consider an artificial data set $D$ with $N = 100$ samples implying *independence* between two binary variables $A$ and $B$, i.e., the empirical distribution factorizes like $\hat{p}(A,B) = \hat{p}(A)\hat{p}(B)$; for simplicity, we assume that $\hat{p}(A) = \hat{p}(B)$. We control the skewness of the marginal distributions with a parameter $z \in [0, 0.5]$: $\hat{p}(A = 1) = z$ and $\hat{p}(A = 0) = 1 - z$; $z = 0.5$ implies uniform distribution, while the skewness of the distribution grows as $z$ decreases. Values in the range $[0.5, 1]$ are symmetric to the case considered here. Figure 1 illustrates how the log Bayes factor (cf. Eq. 7) increases from negative values (implying independence) to positive values (suggesting dependence) as the empirical distribution becomes increasingly skewed (i.e., as $z$ decreases). Figure 1 shows that, as the ESS-value increases, the degree of skewness of $\hat{p}$ has to diminish as to prevent the log Bayes factor from implying dependence. As an aside, note that the curves are quite flat for large and small ESS-value where either one of the distributions dominates the sum; in contrast, the maximum at $z = 0$ is reached for the ESS-value $\alpha = N = 100$, i.e., when both distributions are weighted equally.

---

[2] Because of $N_{a,b,\pi} = N \cdot \hat{p}(a,b,\pi)$.

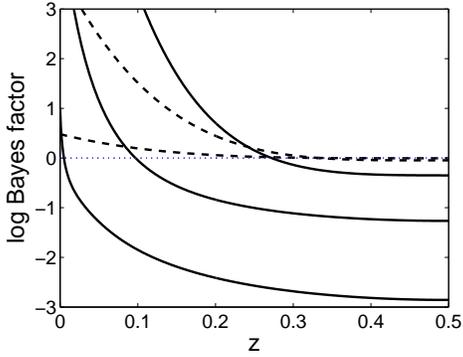

Figure 1: This example illustrates how the dependence increases with the skewness implied by the data for several ESS-values: 100, 10 , 1 (solid curves, from top to bottom); 1,000 and 10,000 (dashed curves, from top to bottom).

## 4 UNDERSTANDING THE OPTIMAL ESS-VALUE

Having theoretically analyzed the cases of large ESS-values in the first part of this paper, and the case of small ESS-values in [Steck & Jaakkola, 2002], the remainder of this paper focuses on the 'optimal' ESS-value (in a predictive sense): we tackle the question as to why the 'optimal' ESS-value is about 50 for some of the 20 UCI data sets in the elaborate experiments in [Silander *et al.*, 2007] (their results are also reproduced in columns $\alpha^M$ and $\alpha^I$ in our Table 1 for comparison), while it ranges between 2 and 13 for the remaining data sets. This section aims to provide an answer to this question, i.e., the goal is to *understand* as to which properties of the data determine the 'optimal' ESS-value.

### 4.1 OPTIMIZATION OF GRAPH AND ESS

Given that the ESS-value can have a crucial effect on the learned network structure [Silander *et al.*, 2007, Steck & Jaakkola, 2002], we now depart from the orthodox Bayesian approach of choosing the prior—including the ESS-value— without having seen the data. In the remainder of this paper we treat the ESS $\alpha$ as a latent random variable to be learned in light of the data $D$. Various objective functions for determining the 'optimal' ESS value $\alpha$ were discussed in [Silander *et al.*, 2007], and found to yield essentially the same result. In this paper, we choose to jointly optimize the graph structure and the ESS-value:

$$(\alpha^*, G^*) = \arg\max_{(\alpha, G)} p(D|\alpha, G) \quad (11)$$

While joint optimization of Eq. 11 is computationally very expensive, a simple coordinate-wise ascent is more tractable. This approach comprises two alternate update-steps in each round $k = 0, 1, ...$ (until convergence):

1. optimize the graph for a fixed ESS-value:
   $G_k^* = \arg\max_G p(D|\alpha_{k-1}^*, G)$,

2. optimize the ESS-value for a fixed graph:
   $\alpha_k^* = \arg\max_\alpha p(D|\alpha, G_k^*)$.

The first step can be solved by any standard algorithm for structure learning (cf. the review in Section 2). Concerning the initialization of the graph $G_0$ in this iterative algorithm, there are several options. A convenient choice is to learn the graph $G_0$ that optimizes the Bayesian Information Criteria (BIC) [Schwarz, 1978, Haughton, 1988]. This criteria not only is a large-sample approximation to the log marginal likelihood in Eq. 2, but it also is independent of ESS, i.e., it does not contain a free parameter. It can hence be used for initialization of $G_0$ when no initial ESS-value is known, and promises to yield an initial graph $G_0$ that is already close to the optimum with respect to the marginal likelihood.

The main contribution of the remainder of this paper is to derive an analytical approximation for the optimal $\alpha^*$ in the second update-step.

### 4.2 OPTIMAL ESS FOR GIVEN GRAPH

This section provides an understanding of the most important properties of the data that affect the value of the optimal ESS. We derive an analytical approximation to the optimal ESS-value for a fixed graph $G$, cf. step 2 above.

Step 2 maximizes predictive accuracy in the prequential sense [Heckerman et al., 1995, Dawid, 1984], when the data points are considered to arrive individually in a sequence. As this optimization problem is difficult to solve, we now replace it by a similar, but *frequentist* objective, departing from *Bayesian* statistics: we minimize the *test error*, as is commonly done in cross-validation. Even though this objective is not the same as the original one, it can be expected to yield a sufficiently accurate approximation in as far as we only aim to understand the difference between ESS-values of about 50 versus about 10 or lower for the various data sets in [Silander et al., 2007]. Note that this assumption and the following ones are validated experimentally on 20 UCI data sets in Section 4.3.

In the following, we combine two approximations to the test error as to obtain an explicit approximation to the optimal ESS-value $\alpha^*$. We carry out both approximations w.r.t. a 'reference point' in the space of distributions. For convenience, we choose this to be the distribution $\hat{p}(X|G)$ implied by the Bayesian network model with graph $G$ and maximum-likelihood parameter estimates $\hat{\theta}$. The test error of this model with respect to the true distribution $p(X)$ reads

$$T[p(X), \hat{p}(X|G)] = -\sum_x p(x) \log \hat{p}(x|G), \quad (12)$$

when using the log loss. As $p$ is unknown, this cannot be evaluated. As is well-known [Akaike, 1973], however, the test error can be approximated by the *training error*, $-E_{\hat{p}(X)}[\log \hat{p}(X|G)]$, and a penalty term for model complexity:

$$T[p(X), \hat{p}(X|G)] = -E_{\hat{p}(X)}[\log \hat{p}(X|G)] + \frac{d_G^{\text{eff}}}{N} + \mathcal{O}(\frac{1}{N^2}), \quad (13)$$

where

$$E_{\hat{p}(X)}[\log \hat{p}(X|G)] = \sum_x \hat{p}(x) \log \hat{p}(x|G)$$
$$= \sum_{i=1}^n \sum_{x_i} \sum_{\pi_i} \frac{N_{x_i,\pi_i}}{N} \log \frac{N_{x_i,\pi_i}}{N_{\pi_i}}, \quad (14)$$

equals the maximum log likelihood divided by the sample size, and $d_G^{\text{eff}}$ is the *effective number of parameters* of the Bayesian network model:

$$d_G^{\text{eff}} = \sum_{i=1}^n \left[ \sum_{x_i,\pi_i} I(N_{x_i,\pi_i}) - \sum_{\pi_i} I(N_{\pi_i}) \right], \quad (15)$$

where $I(\cdot)$ is the indicator function: $I(a) = 1$ if $a > 0$ and $I(a) = 0$ otherwise. If the data set is sufficiently large so that all cell counts $N_{x_i,\pi_i}$ are positive, then $d_G^{\text{eff}}$ equals the well-known number of parameters, which is given for Bayesian networks by $\sum_i (|X_i| - 1)|\Pi_i|$, where $|\cdot|$ denotes the number of (joint) states of the variable(s).

We obtain the second approximation to the test error in Eq. 12 as follows: we assume the (unknown) true distribution $p$ is well-approximated by the functional form of the regularized parameter estimates in Eq. 3 using the optimal value $\alpha^*$:

$$p_{x_i,\pi_i} \approx \tilde{p}_{x_i,\pi_i}, \quad (16)$$

where

$$\tilde{p}_{x_i,\pi_i} = \frac{N_{x_i,\pi_i} + \alpha^* \cdot q_{x_i,\pi_i}}{N + \alpha^*}$$
$$= \hat{p}(x_i, \pi_i) + \frac{\alpha^*}{N}\left(q_{x_i,\pi_i} - \frac{N_{x_i,\pi_i}}{N}\right) + \mathcal{O}((\frac{\alpha^*}{N})^2). \quad (17)$$

This first-order Taylor expansion about the maximum-likelihood estimate (our 'reference point') assumes $\alpha^* \ll N$, which will be justified at the end of this section. Inserting Eqs. 16 and 17 into Eq. 12, now results in our second approximation,

$$T[p(X), \hat{p}(X|G)] \approx -E_{\hat{p}(X)}[\log \hat{p}(X|G)]$$
$$- \frac{\alpha^*}{N} \sum_x (q_x - \frac{N_x}{N}) \log \hat{p}(x|G) + \mathcal{O}\left((\frac{\alpha^*}{N})^2\right)$$

Next we equate this approximation with the one in Eq. 13, and finally arrive at an explicit approximation to the optimal ESS-value,

$$\alpha^* \approx \frac{d_G^{\text{eff}}}{E_{\hat{p}(X)}[\log \hat{p}(X|G)] - E_{q(X)}[\log \hat{p}(X|G)]} + \mathcal{O}(\frac{\alpha^{*2}}{N}), \quad (18)$$

where $E_{q(X)}[\log \hat{p}(X|G)]$ is the expectation with respect to the prior distribution $q$, analogous to Eq. 14; regarding the robustness against zero cell counts, we use $N_{x_i,\pi_i}^+ = \max\{N_{x_i,\pi_i}, 1\}$ in place of $N_{x_i,\pi_i}$ in practice.

Note that the denominator in Eq. 18 is indeed positive if the prior distribution $q$ has a larger entropy $H$ than the empirical distribution $\hat{p}$ does, which is the case for the BDeu score. This is obvious from

$$E_{\hat{p}(X)}[\log \hat{p}(X|G)] - E_{q(X)}[\log \hat{p}(X|G)]$$
$$= H(q(X|G)) - H(\hat{p}(X|G)) + \text{KL}(q(X|G)||\hat{p}(X|G))$$

given that the Kullback-Leibler divergence and the entropy $H$ are non-negative.

**Interpretation of Eq. 18:** This explicit approximation of the optimal ESS $\alpha^*$ provides an understanding of the main properties of the data determining the optimal ESS-value:

- *skewness and dependence*: the denominator of Eq. 18 tends to increase when the entropy (or negative likelihood) of the empirical distribution factored according to the graph structure, $\hat{p}(X|G)$, decreases: this is the cases if there are strong dependencies along the edges, or if the conditional distributions of the variables are very skewed. For such data sets, one can hence expect small values of $\alpha^*$. Conversely, data sets that do neither imply a skewed distribution nor extremely strong dependencies will result in increased values of $\alpha^*$.

- *sample size*: Eq. 18 does not explicitly depend on the sample size $N$. There is an indirect relation, however: the effective number of parameters, which is a measure of model complexity, tends to increase with $N$; a more complex optimal model also entails an increased maximum likelihood, and hence this also affects the denominator; one may hence expect both the enumerator and the denominator to grow in a similar way as $N$ increases, which suggests a weak dependence of $\alpha^*$ on $N$. Apart from that, when $N$ is sufficiently large, the model complexity tends to approach its 'true' value and not increase further. Thus, for a sufficiently large data set, $\alpha^*$ becomes independent of $N$. Consequently, our earlier assumption $\alpha^* \ll N$ indeed holds for sufficiently large $N$. Note that this behavior is also consistent with the one expected in the asymptotic limit ($N \to \infty$), namely $\alpha^*/N \to 0$, so that the effect of regularization vanishes.

- *number of nodes*: Eq. 18 implies that $\alpha^*$ can be expected to be unaffected by the number $n$ of nodes in the (sparse) graph on average because both the enumerator and denominator are additive w.r.t. the nodes, and hence on average grow proportionally to each other when adding additional nodes to the network.

### 4.3 EXPERIMENTS

As an additional confirmation of our assumptions underlying the approximations in Section 4.2, we determined the optimal value $\alpha^*$ based on our iterative algorithm using Eq. 18 on 20 UCI data sets [Hettich & Bay, 1999]. We used the same pre-processed data as was used in [Silander *et al.*, 2007] (imputation of missing values, discretization of continuous variables), and compared to their exact results.

Table 1 summarizes the results, and confirms the validity of our approximation. It is obvious that our approximate $\alpha^*$ agrees very well with the exact results of [Silander *et al.*, 2007], which were obtained under heavy computational costs there (in our Table 1, $\alpha^M$ is the exact solution of the maximization problem in Eq. 11, and $\alpha^I$ is the range of ESS-values that all yield the same MAP graph as $\alpha^M$ does [Silander *et al.*, 2007]): while our approximation $\alpha^*$ does not always agree precisely with the exact value $\alpha^M$, it correctly identifies the data sets where the optimal ESS is about 50 as opposed to about 10 or lower for the remaining data sets. This suggests that our analytical approximation, and the underlying assumptions, capture the main effects that influence the optimal ESS-value. Moreover, note that the sample size $N$ varies between about 100 and 60,000, and the number of variables $n$ ranges from 5 to 15. Table 1 shows that both $N$ and $n$ have no obvious effect on the optimal ESS-value, as expected from our analytical approximation (see discussion in Section 4.2).

Table 1: Experimental validation of our analytical approximations in Section 4.2. See text for details.

| Data | $N$ | $n$ | $\alpha^I$ | $\alpha^M$ | $\alpha^*$ | $k$ |
|---|---|---|---|---|---|---|
| balance | 625 | 5 | 1...100 | 48 | 44 | 1 |
| iris | 150 | 5 | 1...3 | 2 | 2 | 2 |
| thyroid | 215 | 6 | 2...2 | 2 | 3 | 5 |
| liver | 345 | 7 | 3...6 | 4 | 3 | 3 |
| ecoli | 336 | 8 | 7...10 | 8 | 8 | 3 |
| abalone | 4177 | 9 | 6...6 | 6 | 7 | 3 |
| diabetes | 768 | 9 | 3...5 | 4 | 3 | 3 |
| post op | 90 | 9 | 3...5 | 3 | 3 | 3 |
| yeast | 1484 | 9 | 1...6 | 6 | 6 | 2 |
| cancer | 286 | 10 | 6...10 | 8 | 7 | 2 |
| shuttle | 58000 | 10 | 1...3 | 3 | 3 | 2 |
| tictac | 958 | 10 | 51...62 | 51 | 60 | 2 |
| bc wisc | 699 | 11 | 7...15 | 8 | 5 | 3 |
| glass | 214 | 11 | 5...6 | 6 | 6 | 4 |
| page | 5473 | 11 | 3...3 | 3 | 3 | 2 |
| heart cl | 303 | 14 | 13...16 | 13 | 9 | 3 |
| heart hu | 294 | 14 | 5...6 | 5 | 5 | 3 |
| heart st | 270 | 14 | 7...10 | 10 | 10 | 4 |
| wine | 178 | 14 | 8...8 | 8 | 7 | 3 |
| adult | 32561 | 15 | 48...58 | 50 | 49 | 3 |

Moreover, note that the results of our two contributions are very similar to each other (see Sections 3 and 4): the decisive properties of the data are the implied dependencies, or—in case of independence—the skewness of the implied distribution.

With this insight, it is not surprising that an increase in the optimal ESS value is related to a reduction in the maximum number of edges in the graph attainable in the experiments of [Silander *et al.*, 2007]: the data sets with a *large* optimal ESS-value of about 50 are exactly the data sets for which the maximum number of edges in the graph (achieved by increasing the ESS-value) is less than 80% of the edge-count of the complete graph (see the column 'range' in Ta-ble 1 of [Silander *et al.*, 2007]). Both have the same cause, namely a data set that implies neither strong dependencies nor skewness.

As an aside, note that our iterative algorithm (cf. Section 4.1 using the approximation in Eq. 18) is computationally efficient, as it converges within a small number of iterations, cf. $k$ in Table 1; as a convergence criteria we required $|\alpha_k^* - \alpha_{k-1}^*| < 0.1$.

# 5 CONCLUSIONS

This paper presents two contributions that shed light on how the learned graph is affected by the value of the equivalent sample size (ESS). First, we analyzed theoretically the case of *large but finite* ESS values, which complements the results for small values in the literature. Among other results, it was surprising to find that the presence of an edge in a Bayesian network is favoured over its absence even if both the Dirichlet prior and the data imply independence, as long as the conditional empirical distribution is notably different from uniform. Our second contribution provides an understanding of which properties of the given data determine the optimal ESS value in a predictive sense (when considered as a free parameter). Our analytical approximation (which we also validated experimentally) shows that the optimal ESS-value is approximately independent of the number of variables in the (sparse) graph and of the sample size. Moreover, the optimal ESS-value is small if the data implies a skewed distribution or strong dependencies along the edges of the graph. Interestingly, this condition concerning the optimal ESS-value is very similar to the one derived for large but finite ESS-values. Finally, having shown the crucial effect of the ESS value on the graph structure that *maximizes* the Bayesian score, this suggests that a similar effect can be expected concerning the posterior *distribution* over the graphs, and hence for Bayesian model *averaging*. If one embarks on this popular Bayesian approach, one hence has to choose the prior with great care.

#### Acknowledgements

I am grateful to R. Bharat Rao for encouragement and support of this work, to Tomi Silander and Petri Myllymäki for providing me with their pre-processed data, and to the anonymous reviewers for valuable comments.